\documentclass{ieeeaccess}

\usepackage{amssymb}
\usepackage{textcomp}
\usepackage{multirow}
\usepackage{graphicx}
\usepackage{array}
\usepackage{mdwmath}
\usepackage{mdwtab}
\usepackage{eqparbox}
\usepackage{algorithm}
\usepackage{algorithmicx}
\usepackage{algpseudocode}
\usepackage{amsmath}
\usepackage{amsfonts}
\usepackage{url}
\usepackage{amsthm}   
\usepackage{subfigure}  
\usepackage{soul}

\usepackage{xcolor}
\usepackage{pict2e}

\newsavebox{\ORCIDlogo}
\savebox{\ORCIDlogo}{%
\setlength{\unitlength}{\dimexpr 1em/256\relax}%
\begin{picture}(256,256)%
  \color[HTML]{A6CE39}\put(128,128){\circle*{256}}%
  \color{white}%
  \put(78.6,199.2){\circle*{20}}%
  \moveto(70.9,176,9)\lineto(86.3,176,9)\lineto(86.3,69.8)\lineto(70.9,69.8)%
  \closepath\fillpath%
  \moveto(108.9,176.9)\lineto(150.5,176.9)%
  \curveto(190.1,176.9)(207.5,148.6)(207.5 ,123.3)%
  \curveto(207.5,95,8)(186,69.7)(150.7,69.7)%
  \lineto(108.9,69.7)%
  \closepath\fillpath%
  \color[HTML]{A6CE39}%
  \moveto(124.3,83.6)\lineto(148.8,83.6)%
  \curveto(183.7,83.6)(191.7,110.1)(191.7,123.3)%
  \curveto(191.7,144.8)(178,163)(148,163)%
  \lineto(124.3,163)%
  \closepath\fillpath%
\end{picture}%
}

\newcommand\orcidicon[1]{\href{https://orcid.org/#1}{\usebox{\ORCIDlogo}}}

\usepackage[pdfstartview=XYZ,
colorlinks=true,
linkcolor=blue,
urlcolor=blue,
citecolor=blue,
pdftex,
]{hyperref}

\usepackage{caption,setspace}
\usepackage[noadjust]{cite}

\usepackage{mathrsfs}
\usepackage{url}
\usepackage[switch]{lineno}

\def\BibTeX{{\rm B\kern-.05em{\sc i\kern-.025em b}\kern-.08em
    T\kern-.1667em\lower.7ex\hbox{E}\kern-.125emX}}
\begin{document}
\history{}
\doi{}

\title{Soybean Disease Detection via Interpretable Hybrid CNN-GNN: Integrating MobileNetV2 and GraphSAGE with Cross-Modal Attention}
\author{\uppercase{Md Abrar Jahin\authorrefmark{1,2}\orcidicon{0000-0002-1623-3859}, Shahriar Soudeep\authorrefmark{3}\orcidicon{0009-0004-9317-2326}, M. F. Mridha\authorrefmark{3}\orcidicon{0000-0001-5738-1631}, \IEEEmembership{Senior Member, IEEE}, Md. Jakir Hossen\authorrefmark{4}\orcidicon{0000-0002-9978-7987}, \IEEEmembership{Senior Member, IEEE}, and Nilanjan Dey\authorrefmark{5}\orcidicon{0000-0001-8437-498X}, \IEEEmembership{Senior Member, IEEE}}}

\address[1]{Physics and Biology Unit, Okinawa Institute of Science and Technology Graduate University, Okinawa 904-0412, Japan (e-mail: abrar.jahin.2652@gmail.com)}
\address[2]{Thomas Lord Department of Computer Science, Viterbi School of Engineering, University of Southern California, Los Angeles, CA 90007, USA (e-mail: jahin@usc.edu)}
\address[3]{Department of Computer Science, American International University-Bangladesh, Dhaka 1229, Bangladesh (e-mail: s.shahriar32322323@gmail.com, firoz.mridha@aiub.edu)}
\address[4]{Department of Robotics and Automation, Multimedia University, Melaka, Malaysia (e-mail: jakir.hossen@mmu.edu.my)}
\address[5]{Department of Computer Science \& Engineering, Techno International New Town, Kolkata 700156, India (e-mail: nilanjan.dey@tint.edu.in)}

\markboth
{Jahin \headeretal: Soybean Disease Detection via Interpretable Hybrid CNN-GNN}
{Jahin \headeretal: Soybean Disease Detection via Interpretable Hybrid CNN-GNN}

\corresp{Corresponding author: Md. Jakir Hossen(e-mail:akir.hossen@mmu.edu.my) and M. F. Mridha (e-mail: firoz.mridha@aiub.edu).}

\begin{abstract}
Soybean leaf disease detection is critical for agricultural productivity but faces challenges due to visually similar symptoms and limited interpretability of conventional methods. While convolutional neural networks (CNNs) excel in spatial feature extraction, they often neglect inter-image relational dependencies, leading to misclassifications. This paper proposes an interpretable hybrid sequential CNN-Graph Neural Network (GNN) framework that synergizes MobileNetV2 for localized feature extraction and GraphSAGE for relational modeling. The framework constructs a graph where nodes represent leaf images, with edges defined by cosine similarity-based adjacency matrices and adaptive neighborhood sampling. This design captures fine-grained lesion features and global symptom patterns, addressing inter-class similarity challenges. Cross-modal interpretability is achieved via Grad-CAM and Eigen-CAM visualizations, generating heatmaps to highlight disease-influential regions. Evaluated on a dataset of ten soybean leaf diseases, the model achieves 97.16\% accuracy, surpassing standalone CNNs ($\le$95.04\%) and traditional machine learning models ($\le$77.05\%). Ablation studies validate the superiority of the sequential architecture over parallel or single-model configurations. With only 2.3 million parameters, the lightweight MobileNetV2-GraphSAGE combination ensures computational efficiency, enabling real-time deployment in resource-constrained environments. The proposed approach bridges the gap between accurate classification and practical applicability, offering a robust, interpretable tool for agricultural diagnostics while advancing CNN-GNN integration in plant pathology research.
\end{abstract}


\titlepgskip=-15pt

\maketitle

\section{Introduction}
\label{sec:introduction}
\PARstart{S}{oybean} (\textit{Glycine max}) is one of the most significant crops worldwide, providing essential nutrients and oil for both human consumption and animal feed. However, various diseases, including soybean rust, Septoria brown spot, and frog eye leaf spot, often threaten its production. These diseases severely affect the quality and yield of soybean crops, leading to substantial economic losses for farmers. Traditional methods of disease detection, which are primarily based on manual inspection, are time-consuming, labor-intensive, and subjective, making them unsuitable for large-scale, automated applications.

With the advent of machine learning and deep learning, significant progress has been made in automating plant disease detection, particularly through convolutional neural networks (CNNs). CNNs have demonstrated strong performance in image classification tasks by automatically learning spatial features from raw images, eliminating the need for manual feature extraction. Numerous studies have highlighted their effectiveness in soybean leaf disease classification \cite{chen_using_2020,sethy_deep_2020,dou_classification_2023,Sheng,Bera,Rahman,Janarthan,Wang_2025,Wu_2024}. However, most existing approaches — whether CNNs or transfer learning techniques are used \cite{KARLEKAR_2020,wu2023classification} — focus on extracting features from individual images, overlooking critical relational information between images. This becomes particularly problematic when diseases present visually similar symptoms triggered by different factors, such as nutrient deficiencies, pest damage, or environmental stress, often leading to misclassifications. Moreover, these conventional models offer limited explainability, providing little insight into which leaf regions drive predictions and reducing interpretability and trust among agricultural experts.

To address these limitations, graph neural networks (GNNs) have emerged as a complementary approach capable of modeling relational dependencies between samples. GNNs are particularly well suited for cases where relationships between images, such as symptom similarity or shared environmental conditions, provide valuable diagnostic cues \cite{Senthil2023,Li2024}. By treating images as nodes and defining edges on the basis of pairwise similarities, GNNs aggregate information from neighboring images, enabling context-aware classification incorporating local features and global relational patterns. However, GNNs alone lack the ability to extract fine-grained spatial features directly from raw images — a key strength of CNNs. Therefore, combining CNNs and GNNs into a hybrid framework offers a synergistic advantage: CNNs capture localized spatial features within individual images, whereas GNNs enrich these representations with relational context across images. This hybrid approach is particularly valuable for soybean leaf disease classification, where local lesion characteristics and broader symptom similarity across fields, varieties, and conditions are essential for accurate and interpretable diagnosis.

To address these gaps, we propose an interpretable hybrid sequential CNN-GNN architecture that sequentially combines MobileNetV2 for efficient spatial feature extraction and graph sample and aggregation (GraphSAGE), a GNN architecture, for relational dependency modeling between soybean leaf images. By constructing a similarity graph where nodes represent leaf images and edges encode pairwise feature similarity, GraphSage \cite{hamilton_inductive_2017} aggregates information from neighboring nodes, enriching the feature representations with a relational context. This fusion of local spatial learning and global relational learning enhances classification accuracy while ensuring computational efficiency, making the model suitable for real-time field deployment. Additionally, we incorporate Grad-CAM and Eigen-CAM visualizations to provide interpretable heatmaps that highlight the specific leaf regions influencing each classification decision, bridging the gap between model predictions and expert validation. To the best of our knowledge, this is the first interpretable CNN-GNN hybrid framework applied to soybean leaf disease detection, addressing critical gaps in relational modeling, model transparency, and computational efficiency in plant disease classification research.

We make the following key contributions in this work:
\begin{enumerate} 
\item \textbf{Sequential CNN-GNN Architecture:} We propose a novel pipeline combining a pretrained MobileNetV2 for local feature extraction and a GraphSAGE model for global relational reasoning, enhancing our model’s ability to capture fine-grained disease symptoms and inter-symptom dependencies. 
\item \textbf{Graph Construction with Node Fusion and Adaptive Sampling:} We introduce a domain-specific graph construction where each image is represented as a node with embeddings that fuse spatial and semantic features, whereas adaptive neighborhood sampling ensures robust classification even with similar symptoms or background noise.
\item \textbf{Cross-Modal Interpretability:} We employ Grad-CAM and Eigen-CAM for both CNN and graph-level feature attribution, providing clear insights into which image regions and relational cues contributed to our model’s decision and enhancing transparency in disease diagnosis.
\end{enumerate}

The remainder of this paper is organized as follows: Section \ref{sec2} reviews related work in plant disease detection via CNNs and GNNs. Section \ref{sec3} describes the proposed methodology, including the model architecture. Section \ref{sec4} presents the data preprocessing and experimental setup. Section \ref{sec5} discusses the results and compares the performance of the proposed model with other baseline models, and Section \ref{sec6} concludes the paper with suggestions for future research.

\section{Literature Review}\label{sec2}
Soybean disease identification has become a key research focus in smart agriculture, with machine learning and deep learning techniques significantly improving classification accuracy. Early methods relied on traditional image processing and handcrafted features, such as K-means clustering and SVMs~\cite{padol_svm_2016} or Gabor filters with ANNs~\cite{kulkarni_applying_2012}, but these approaches struggled to generalize across diverse symptoms and complex backgrounds.

Recent agricultural image classification research has increasingly adopted deep learning, which has demonstrated strong performance across various crops and datasets. Chen \emph{et al.}~\cite{chen_using_2020} introduced LeafNetCNN, which achieved 90.16\% accuracy for tea plant diseases, whereas Sethy \emph{et al.}~\cite{sethy_deep_2020} combined CNN feature extraction with SVM classification for rice leaf diseases. Dou \emph{et al.}~\cite{dou_classification_2023} achieved 98.75\% accuracy in citrus disease classification using a CBAM-MobileNetV2 model. Lightweight and attention-based models, such as GSNet~\cite{Sheng}, RAFA-Net~\cite{Bera}, and LiRAN~\cite{Janarthan}, have further enhanced classification accuracy while reducing computational complexity. Studies by Rahman \emph{et al.}~\cite{Rahman} and Wang \emph{et al.}\cite{Wang_2025} also highlight the benefits of tailored CNN architectures for crop and pest classification, addressing challenges such as background complexity and inter-class similarity. Wu \emph{et al.}\cite{Wu_2024} introduced ResNet9-SE, which achieved 99.7\% accuracy in strawberry disease detection by incorporating squeeze-and-excitation blocks. Hyperspectral imaging has also been explored for plant disease detection, offering rich spectral information but introducing significant computational challenges~\cite{Rayhana_2023}.

GNNs have emerged as effective tools for capturing relational dependencies, particularly in domains where the contextual similarity between samples influences classification outcomes. While CNNs excel at extracting spatial features within individual images, they lack the ability to model relationships between samples—an essential capability for tasks such as disease diagnosis, where symptoms may appear subtly across different conditions. Kipf and Welling \cite{kipf_semi-supervised_2017} introduced the foundational graph convolutional network (GCN), which enables node classification through spectral graph convolution, whereas GraphSAGE \cite{Hamilton_graphsage} extended this method to inductive learning, making it suitable for evolving datasets such as agricultural image collections. GNNs have since shown strong performance in medical diagnosis \cite{ahmedt-aristizabal_graph-based_2021}, remote sensing \cite{Kavran_2023}, and plant disease detection \cite{Senthil2023,Li2024}, where capturing both local and global structural dependencies enhances classification accuracy.

Hybrid CNN-GNN architectures have emerged as promising solutions to combine the strengths of spatial feature extraction and relational modeling. Thangamariappan \emph{et al.} \cite{Thangamariappan_2024} demonstrated that integrating CNN-extracted spatial features with GNN-derived relational embeddings improves classification for structured images. Similarly, Nikolentzos \emph{et al.} \cite{nikolentzos_2021} proposed converting images into graphs, where nodes represent pixels or segments and edges capture spatial or semantic proximity, enabling explicit relational modeling. Hua and Li \cite{zhang_2024} applied a multiscale attention-enhanced CNN-GNN framework to remote sensing change detection, which demonstrated improved spatial coherence and detection accuracy. Tang \emph{et al.} \cite{tang_2022} further optimized graph construction to reduce computational overhead while preserving classification performance. While these studies highlight the potential of CNN-GNN hybrids, they focus largely on structured imagery, leaving agricultural disease classification, where symptoms are often subtle, variable, and environment-dependent, relatively underexplored.

In the domain of soybean leaf disease detection, current research has focused largely on standalone CNN models and transfer learning techniques, with limited attention given to relational modeling. Existing studies \cite{KARLEKAR_2020,wu2023classification} primarily employ deep CNN architectures trained directly on leaf images, which achieve reasonable accuracy but lack mechanisms to capture inter-image relationships that could improve robustness, especially in cases where different diseases present visually similar symptoms. Furthermore, most existing methods lack interpretability, providing little insight into the specific features or regions driving predictions, which reduces confidence and usability for domain experts such as plant pathologists and agronomists.

\begin{figure*}[!ht]
\centering
\includegraphics[width=\linewidth]{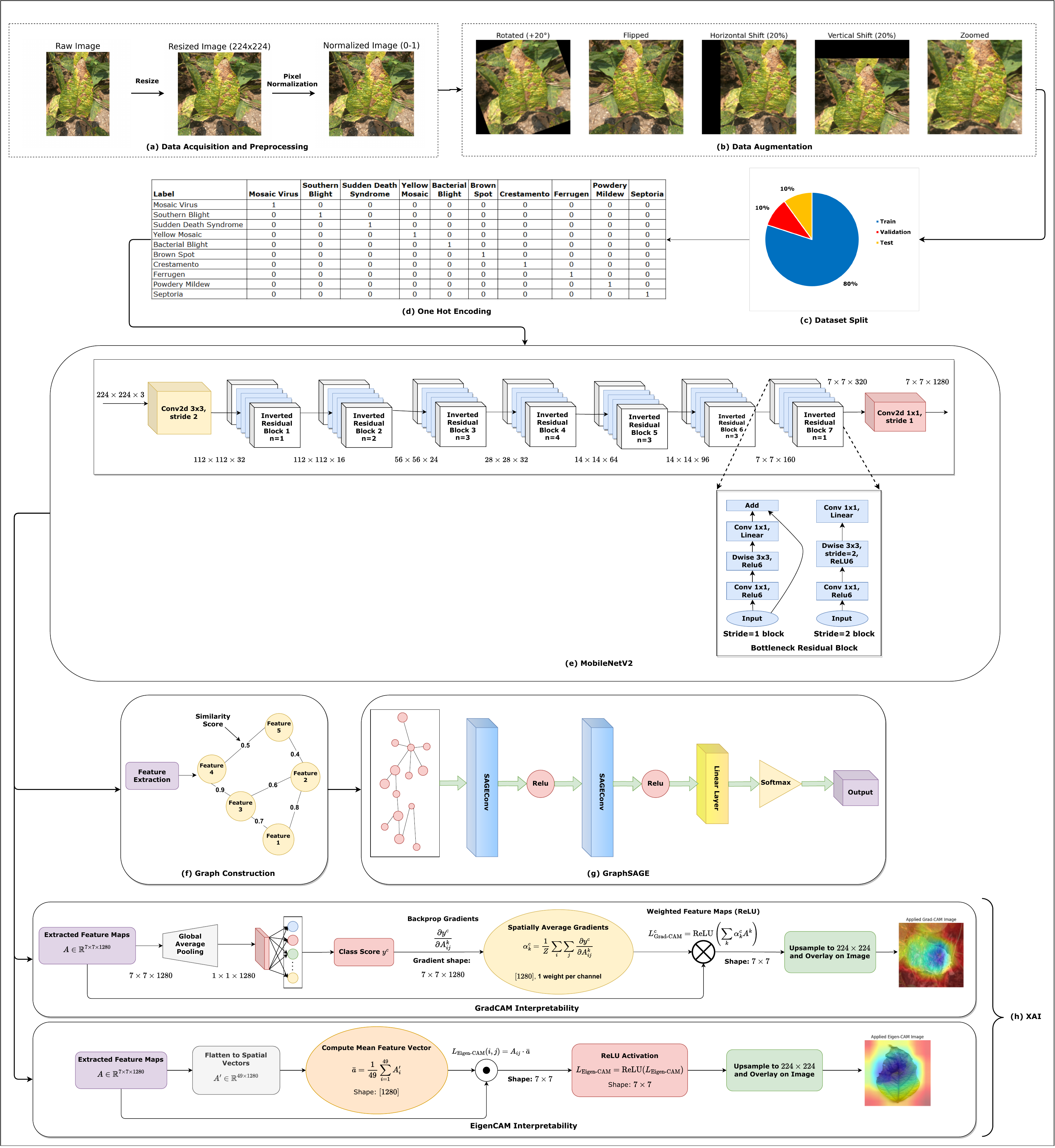}
\caption{Proposed Sequential MobileNetV2-GraphSAGE framework: (a) Images are resized to \(224 \times 224\) pixels and normalized. (b) Data augmentation includes rotation, flipping, shifting, and zooming. (c) Dataset is split (80\%-10\%-10\%) for training, validation, and testing. (d) Disease labels are one-hot encoded. (e) MobileNetV2 extracts local features. (f) Graph construction captures relationships. (g) GraphSAGE aggregates neighborhood information. (h) Cross-modal interpretation uses Grad-CAM and Eigen-CAM.}
\label{fig:framework}
\end{figure*}

\section{Methodology}\label{sec3}

This section outlines the methodology used in this work for classifying soybean leaf diseases via a sequential CNN-GNN model. The overall pipeline, consisting of data preprocessing, augmentation, and model development used in this study, is illustrated in Figure~\ref{fig:framework}. The pipeline consists of several key steps, including image acquisition, resizing, pixel normalization, augmentation, dataset splitting, and one-hot encoding, ensuring a standardized dataset for training and evaluation. During model development, images are first passed through a MobileNetV2 architecture to extract local features. These features are then structured into a graph, with edges representing the relationships between image features within the data. Finally, the aggregated features are fed into a classifier to predict soybean leaf diseases. This integrated approach ensures local feature extraction and global dependency modeling, leveraging the strengths of both the CNN and GNN architectures for improved disease classification.

\subsection{Model Framework and Architecture}
The proposed Sequential CNN-GNN model combines a CNN for extracting local features from images and a GNN for capturing the relationships between these features. The model is designed sequentially: the CNN first extracts detailed local features, and then the GNN processes these features to understand global dependencies, enhancing classification accuracy.

The model begins by taking an input image \( I \in \mathbb{R}^{H \times W \times C} \), where \( H \), \( W \), and \( C \) represent the height, width, and number of channels (3 for RGB), respectively. The input image is first passed through the MobileNetV2 CNN, which is known for its efficiency and low computational cost. Compared with traditional convolutions, MobileNetV2 uses depthwise separable convolutions, which reduce the number of parameters and operations. This makes it ideal for extracting local features efficiently.

The output from the MobileNetV2 CNN is a set of feature maps \( F_{\text{cnn}} \in \mathbb{R}^{H' \times W' \times C'} \), where \( H' \), \( W' \), and \( C' \) represent the spatial dimensions and depth of the extracted feature maps. These feature maps are then normalized to a range between 0 and 1 by dividing by 255, ensuring consistent scaling across the dataset:
\begin{equation}
F_{\text{cnn}} = \frac{F_{\text{cnn}}}{255}
\end{equation}

These normalized feature maps serve as inputs for the GNN branch, where each feature map is treated as a node in a graph. To construct the graph, we calculate the cosine similarity between feature vectors from different image patches to measure their relationships. The similarity between two feature vectors \( F_i \) and \( F_j \) is computed as:
\begin{equation}
S_{ij} = \frac{F_i \cdot F_j}{\| F_i \| \| F_j \|}
\end{equation}
where \( \| F_i \| \) and \( \| F_j \| \) are the L2 norms of \( F_i \) and \( F_j \), respectively. An adjacency matrix \( A \) is constructed on the basis of these similarities. If the similarity between two nodes exceeds a threshold \( \theta \), we set \( A_{ij} = 1 \); otherwise, \( A_{ij} = 0 \). This adjacency matrix defines how the nodes (image patches) are connected in the graph.

GraphSAGE is then used to aggregate information from the neighbors of each node. Unlike traditional GNNs, which aggregate information from all neighboring nodes, GraphSAGE performs neighborhood sampling to handle large graphs efficiently. The feature update rule for a node \( H^{(k+1)} \) at the \( (k+1) \)-th layer is as follows:
\begin{equation}
H^{(k+1)} = \sigma \left( \hat{A} H^{(k)} W^{(k)} \right)
\end{equation}
where \( H^{(k)} \in \mathbb{R}^{N \times F_k} \) is the feature matrix at the \( k \)-th layer, with \( N \) nodes and \( F_k \) features per node. \( \hat{A} \) is the normalized adjacency matrix (with self-loops added), \( W^{(k)} \in \mathbb{R}^{F_k \times F_{k+1}} \) is the learnable weight matrix at layer \( k \), and \( \sigma \) is the nonlinear activation function (typically ReLU).

GraphSAGE aggregates the features of neighboring nodes via a sampling-based approach and iteratively updates the node features to capture both local and global information. After several layers of graph convolutions, the final node features \( H^{(K)} \) are passed through a Softmax layer to compute the class probabilities:
\begin{equation}
\hat{y} = \text{softmax}(H^{(K)})
\end{equation}
These probabilities represent the likelihood of each class for the image.

\subsubsection{Sequential Architecture}
The architecture of the sequential CNN-GNN model combines the strengths of both CNNs and GNNs. The CNN branch captures local image patterns, whereas the GraphSAGE-based GNN branch aggregates information from neighboring patches to understand global relationships. This sequential structure allows the model to leverage local and global feature representations, making it particularly effective for complex image classification tasks, such as soybean leaf disease identification.
\begin{enumerate}
    \item \textbf{Input:} The model accepts an image \( I \in \mathbb{R}^{H \times W \times C} \).
    \item \textbf{CNN Branch:} MobileNetV2 extracts local features, resulting in feature maps \( F_{\text{cnn}} \in \mathbb{R}^{H' \times W' \times C'} \).
    \item \textbf{GNN Branch:} The feature maps are treated as nodes in a graph. The adjacency matrix \( A \) is constructed via cosine similarity, and the GraphSAGE algorithm updates the node features via graph convolutions.
    \item \textbf{Output:} The final node features are passed through a Softmax layer to output class probabilities.
\end{enumerate}

\subsection{Feature Extraction Techniques}
The feature extraction step leverages MobileNetV2, a lightweight convolutional neural network architecture specifically designed for efficient image classification tasks, particularly in resource-constrained environments. MobileNetV2 uses depthwise separable convolutions, significantly reducing computational complexity and the number of parameters while maintaining high performance. This makes MobileNetV2 especially effective for extracting meaningful local features from images without incurring high computational costs, which is crucial for large-scale agricultural datasets.

For the soybean leaf disease classification task, MobileNetV2 effectively captures local image features, which are essential for distinguishing between different leaf disease types. Mathematically, the feature extraction process can be represented as:
\begin{equation}
F_{\text{cnn}} = \text{MobileNetV2}(I)
\end{equation}
where \( F_{\text{cnn}} \in \mathbb{R}^{H' \times W' \times C'} \) represents the feature map extracted from the input image \( I \), with \( H' \), \( W' \), and \( C' \) denoting the spatial dimensions and depth of the extracted features.

After extraction, the feature maps undergo normalization to ensure that the pixel values fall within the range [0, 1]. This normalization step standardizes the subsequent GNN input, ensuring consistent scaling across the dataset. This consistency is crucial for improving model convergence during training and helping the model learn more effectively.

\subsection{Graph Construction and Representation}
After feature extraction, each feature map \( F_{\text{cnn}} \) is treated as a node in the graph. The relationship between these nodes is modeled by computing the cosine similarity between the feature vectors extracted from different image patches. This similarity helps capture the structural and semantic relationships between the image regions, which is critical for understanding global dependencies.

The adjacency matrix \( A \) is then constructed on the basis of these cosine similarities. Each element \( A_{ij} \) in the matrix represents the relationship between nodes \( i \) and \( j \), with higher values indicating a stronger relationship. To simplify the graph structure, we threshold the cosine similarity to form a binary adjacency matrix:
\begin{equation}
A_{ij} =
\begin{cases}
1 & \text{if } S_{ij} > \theta \\
0 & \text{otherwise}
\end{cases}
\end{equation}
where \( S_{ij} \) denotes the cosine similarity between nodes \( i \) and \( j \) and where \( \theta \) is the similarity threshold. The threshold \( \theta \) controls how strongly nodes must be related to be connected in the graph. This binary representation helps reduce complexity while preserving the most meaningful relationships. This graph structure enables the GNN to model and learn the interdependencies between image patches, capturing local and global patterns for enhanced classification performance.

\subsection{Optimization Methods and Loss Function}
The model is trained via categorical cross-entropy as the loss function, which is suitable for multi-class classification. The loss function is defined as:
\begin{equation}
L = - \sum_{i} y_i \log(\hat{y_i})
\end{equation}
where \( y_i \) is the true class label for the \( i \)-th image and \( \hat{y_i} \) is the predicted probability for class \( i \).

We use the Adam optimizer with a learning rate of 0.001 for model optimization, as it is efficient and adapts the learning rate during training. Dropout is applied to fully connected layers to prevent overfitting, with a dropout rate of 0.5.

\subsection{Model Training and Evaluation}
The model was trained via an 80\%-20\% data split, with 80\% of the dataset allocated for training and 20\% allocated for testing. A batch size of 32 was used, and training was conducted for 20 epochs to ensure sufficient learning. The model's performance was evaluated via standard classification metrics, including accuracy, precision, recall, and F1 score. Accuracy represents the percentage of correctly classified samples, whereas precision measures the proportion of correctly predicted positive instances out of all predicted positives. Recall quantifies the model’s ability to identify all actual positive cases, and the F1 score provides a balanced measure by computing the harmonic mean of precision and recall.

\subsection{Mathematical Formulations}
GraphSAGE updates node features through a neighborhood sampling and aggregation process designed to efficiently handle large-scale graphs. Instead of directly using the entire adjacency matrix, GraphSAGE samples a fixed-size set of neighboring nodes for each target node at every layer. For a given node \(v\) at layer \(k\), its feature representation is updated by aggregating the feature vectors of its sampled neighbors. This aggregation can be performed via different strategies, such as the mean aggregation, pooling, or an LSTM-based aggregator. The aggregated neighbor features are then concatenated with the current node’s own features, and the concatenated vector is passed through a learnable linear transformation followed by a nonlinear activation function (typically ReLU). Mathematically, the feature update at layer \(k\) can be expressed as:
\begin{equation}
h_v^{(k+1)} = \sigma \left(W^{(k)} \cdot \text{AGGREGATE}\left(\{h_u^{(k)} : u \in \mathcal{N}(v)\}\right) \, \Vert \, h_v^{(k)}\right)
\end{equation}
where \(h_v^{(k)}\) denotes the feature vector of node \(v\) at layer \(k\), \(\mathcal{N}(v)\) denotes the sampled neighborhood of node \(v\), \(W^{(k)}\) is a trainable weight matrix, \(\sigma\) is a nonlinear activation function, and \(\Vert\) denotes the concatenation operation. This sampling and aggregation process allows GraphSAGE to scale efficiently to large graphs while maintaining flexibility in how neighbor information is combined. After \(K\) layers of neighborhood aggregation, the final node embeddings \(h_v^{(K)}\) can be directly used for downstream tasks such as node classification, where they are passed through a Softmax layer to compute class probabilities:
\begin{equation}
\hat{y}_v = \text{softmax}(h_v^{(K)})
\end{equation}
This formulation allows GraphSAGE to learn expressive node representations while being computationally efficient, as it avoids the need to process all neighbors at every step, unlike traditional GCNs do.

\subsection{Innovative Techniques}
The proposed model leverages a hybrid architecture that combines MobileNetV2 with GraphSAGE, capitalizing on the complementary strengths of CNNs and GNNs. MobileNetV2 is a lightweight feature extractor that efficiently captures fine-grained local patterns from the input image through depthwise separable convolutions. These extracted features are then transformed into graph-structured data, where GraphSAGE aggregates information from neighboring nodes, enabling the model to learn spatial and semantic relationships between localized regions. This explicit modeling of local texture details and global relational dependencies enhances the model’s ability to distinguish subtle variations between disease patterns, particularly in cases where visual symptoms exhibit spatial spread or irregular clustering. Compared with traditional CNN pipelines, this hybrid design reduces reliance on large convolutional stacks, improving computational efficiency while enhancing spatial reasoning — a limitation in purely convolutional architectures. Moreover, unlike GCN and GAT, which assume fixed graph structures or require dense attention computations, GraphSAGE’s sampling-based neighborhood aggregation balances performance and scalability, making it well suited for irregular and incomplete spatial patterns common in leaf disease imaging. This combination of efficient feature extraction, flexible graph modeling, and reduced computational overhead positions the proposed approach as a robust alternative to standalone CNNs, standalone GNNs, parallel CNN-GNN connections, and other hybrid CNN-GNN variants.

\section{Experiments}\label{sec4}

\subsection{Datasets and Preprocessing}
The dataset\footnote{\url{https://www.kaggle.com/datasets/sivm205/soybean-diseased-leaf-dataset}} utilized in this study comprises high-quality images of soybean leaves affected by ten different diseases, including mosaic virus, southern blight, sudden death syndrome, yellow mosaic, bacterial blight, brown spot, Crestamento, Ferrugen, powdery mildew, and Septoria. The dataset is well labelled and encompasses a diverse range of real-world conditions, making it highly suitable for plant disease classification tasks. To ensure consistency in the input size, all the images were resized to $224\times224$ pixels. Pixel values were normalized by scaling them between 0 and 1, which was achieved by dividing by 255. Data augmentation techniques, such as random rotation up to $20^\circ$, horizontal flipping, width and height shifts of 20\%, and zooming, were applied to increase model robustness. The dataset was partitioned into 80\% training and 20\% testing, with 10\% of the total dataset reserved for validation during training, while the remaining 10\% was utilized for final testing. Furthermore, the categorical disease labels were one-hot encoded to facilitate multi-class classification. These preprocessing steps ensured that the model was trained on standardized and augmented data, improving its ability to generalize effectively to unseen samples.

\subsection{Experimental Configuration}  
The experiments were conducted via a GPU $T4\times2$ setup to enable efficient and accelerated training. The deep learning framework utilized for model development was TensorFlow 2.x with Keras, along with essential libraries such as NumPy, Matplotlib, and Scikit-learn. The implementation was performed in Python 3.8. The training process employed the Adam optimizer with a learning rate of 0.001, a batch size of 32, and 20 epochs. To mitigate overfitting, a dropout rate of 0.5 was applied to the fully connected layers. The model's performance was evaluated in terms of accuracy, precision, recall, and F1 score. To assess the effectiveness of the sequential CNN-GNN model, we conducted comparisons against baseline models and performed ablation tests.

\subsection{Evaluation Strategy}
To demonstrate the benefits of integrating local feature extraction from CNNs with global relational reasoning from GNNs, we compare the proposed model against a diverse set of baselines, including traditional machine learning models (support vector classifiers, random forests, logistic regression, K-nearest neighbors), standalone CNNs \textit{(MobileNetV2, EfficientNetB0, ResNet50, VGG16, VGG19, Xception, DenseNet family, InceptionV3, NASNetLarge, and ResNet variants)}, and hybrid combinations of CNNs and GNNs \textit{(GCN, GAT, and GraphSAGE)}. This comprehensive evaluation highlights the strengths and limitations of each architecture, particularly the ability of the MobileNetV2-GraphSAGE combination to balance lightweight feature extraction with neighborhood-aware reasoning. To ensure fairness and reproducibility, all the models were trained and evaluated under the same conditions, using identical batch sizes, learning rates, optimizers, number of epochs, data augmentation pipelines, and training-validation-test splits. This consistent experimental setup ensures that performance differences reflect genuine architectural advantages rather than variations in training protocols.

\section{Results and Discussion}\label{sec5}

\subsection{Performance Comparison}
The benchmarking results, as shown in Table \ref{tab:benchmarking}, provide critical insights into the performance of various CNN-GNN hybrid models for soybean leaf disease classification. Among the tested models, MobileNetV2 + GraphSAGE and InceptionV3 + GraphSAGE achieved the highest accuracy of 97.16\%, with MobileNetV2 + GraphSAGE showing a slight edge in precision (97.51\%) and InceptionV3 + GraphSAGE excelling in the F1 score (97.06\%). These results indicate that integrating GraphSAGE with lightweight CNN architectures can significantly enhance classification performance. 
On the other hand, traditional machine learning models such as SVC, random forest, logistic regression, and KNN showed significantly lower performance, with accuracies ranging from 66.39\% to 77.05\%. These results highlight the superiority of deep learning models, particularly the sequential CNN-GNN, over traditional machine learning models for this image classification task.

GraphSAGE consistently outperformed the GCN and GAT across all the CNN backbones, highlighting its superior ability to extract meaningful graph-based features. While deeper CNN architectures such as DenseNet201 and DenseNet169 also demonstrated strong results, achieving accuracy above 96\%, their performance was slightly lower than that of the best-performing models. EfficientNetB0 exhibited extremely poor results ($\approx$15.6\% accuracy) across all the GNN variants, indicating its inefficacy in this classification task.

Furthermore, ResNet101 and ResNet152 performed poorly with GCN but improved significantly with GraphSAGE and GAT, emphasizing the importance of selecting the right GNN variant for a given CNN backbone. Overall, the results confirm that combining lightweight CNNs with GraphSAGE is the most effective approach for soybean leaf disease classification. Among the models, MobileNetV2 + GraphSAGE stands out as the proposed model because of its superior balance between accuracy, computational efficiency, and ease of deployment. While InceptionV3 + GraphSAGE achieved similar accuracy, MobileNetV2’s lightweight architecture makes it more suitable for real-world applications, particularly in resource-constrained environments where efficiency and scalability are critical.

According to Table~\ref{tab:param}, traditional CNNs such as ResNet, VGG, Inception, and NASNet have larger parameter counts, with NASNetLarge at 84.9 M. These models are more complex and resource-intensive. MobileNetV2 and EfficientNetB0 are lightweight models with 2-4 M parameters, which are ideal for mobile or edge devices. DenseNet models (7 M to 18 M parameters) feature dense connectivity for better information flow with a moderate size. GCN, GAT, and GraphSAGE, with 10K--67K parameters, are specialized for graph tasks and have many fewer parameters than CNNs do. Larger models such as InceptionV3 and ResNet152 offer high performance but require more resources, whereas smaller models such as MobileNetV2 balance performance and efficiency.

\begin{table}[!ht]
\centering
\caption{Benchmarking Results on the Soybean Leaf Disease Dataset. \textbf{Bold} indicates the best performance and \ul{Underline} indicates the second best performance.}
\label{tab:benchmarking}
\resizebox{\columnwidth}{!}{%
\begin{tabular}{lcccc}
\hline
\textbf{Model} & \textbf{Accuracy ($\uparrow$)} & \textbf{Precision ($\uparrow$)} & \textbf{Recall ($\uparrow$)} & \textbf{F1 Score ($\uparrow$)} \\
\hline
SVC & 77.05\% & 76.00\% & 77.05\% & 73.65\% \\
RandomForest & 77.05\% & 74.86\% & 77.05\% & 73.65\% \\
Logistic Regression & 77.05\% & 73.39\% & 77.05\% & 74.42\% \\
KNN & 66.39\% & 70.00\% & 66.39\% & 61.21\% \\
MobileNetV2         & 93.62\% & 93.75\% & 93.62\% & 92.77\% \\
MobileNetV2 + GCN & 95.74\% & 95.60\% & 95.74\% & 95.46\% \\
MobileNetV2 + GAT & \ul{96.45\%} & 96.89\% & \ul{96.45\%} & 96.11\% \\
\textbf{MobileNetV2 + GraphSAGE} & \textbf{97.16\%} & \textbf{97.51\%} & \textbf{97.16\%} & \ul{96.79\%} \\
EfficientNetB0      & 19.86\% & 3.94\%  & 19.86\% & 6.58\% \\
EfficientNetB0 + GCN & 15.60\% & 2.43\% & 15.60\% & 4.21\% \\
EfficientNetB0 + GAT & 15.60\% & 2.43\% & 15.60\% & 4.21\% \\
EfficientNetB0 + GraphSAGE & 15.60\% & 2.43\% & 15.60\% & 4.21\% \\
ResNet50            & 47.52\% & 35.18\% & 47.52\% & 36.20\% \\
ResNet50 + GCN & 50.35\% & 45.06\% & 50.35\% & 43.79\% \\
ResNet50 + GAT & 62.41\% & 61.20\% & 62.41\% & 57.48\% \\
ResNet50 + GraphSAGE & 63.12\% & 63.81\% & 63.12\% & 58.90\% \\
VGG16                & 82.98\% & 80.23\% & 82.98\% & 79.29\% \\
VGG16 + GCN & 94.33\% & 91.88\% & 94.33\% & 93.03\% \\
VGG16 + GAT & 94.33\% & 92.62\% & 94.33\% & 93.36\% \\
VGG16 + GraphSAGE & 93.62\% & 91.44\% & 93.62\% & 92.38\% \\
VGG19                & 80.14\% & 78.19\% & 80.14\% & 76.14\% \\
VGG19 + GCN & 95.04\% & 91.97\% & 95.04\% & 93.40\% \\
VGG19 + GAT & 92.20\% & 90.03\% & 92.20\% & 90.78\% \\
VGG19 + GraphSAGE & 95.04\% & 91.97\% & 95.04\% & 93.40\% \\
Xception             & 92.20\% & 92.78\% & 92.20\% & 91.20\% \\
Xception + GCN & 94.33\% & 94.81\% & 94.33\% & 94.34\% \\
Xception + GAT & 95.04\% & 95.26\% & 95.04\% & 94.96\% \\
Xception + GraphSAGE & 95.04\% & 95.13\% & 95.04\% & 94.95\% \\
DenseNet121         & 92.91\% & 92.76\% & 92.91\% & 92.37\% \\
DenseNet121 + GCN & 93.62\% & 91.69\% & 93.62\% & 92.45\% \\
DenseNet121 + GAT & 95.04\% & 95.65\% & 95.04\% & 94.70\% \\
DenseNet121 + GraphSAGE & 95.74\% & 95.45\% & 95.74\% & 95.45\% \\
DenseNet169         & 92.91\% & 90.43\% & 92.91\% & 91.15\% \\
DenseNet169 + GCN & \ul{96.45\%} & 96.11\% & \ul{96.45\%} & 96.14\% \\
DenseNet169 + GAT & 95.74\% & 94.96\% & 95.74\% & 95.27\% \\
DenseNet169 + GraphSAGE & \ul{96.45\%} & 96.73\% & \ul{96.45\%} & 95.59\% \\
DenseNet201         & 95.04\% & 93.03\% & 95.04\% & 93.75\% \\
DenseNet201 + GCN & 94.33\% & 94.17\% & 94.33\% & 93.82\% \\
DenseNet201 + GAT & \textbf{97.16\%} & 96.87\% & \textbf{97.16\%} & 96.87\% \\
DenseNet201 + GraphSAGE & \ul{96.45\%} & 96.32\% & \ul{96.45\%} & 96.20\% \\
InceptionV3         & 92.20\% & 92.10\% & 92.20\% & 91.73\% \\
InceptionV3 + GCN & \ul{96.45\%} & 96.70\% & \ul{96.45\%} & 96.05\% \\
InceptionV3 + GAT & 95.04\% & 95.55\% & 95.04\% & 94.21\% \\
InceptionV3 + GraphSAGE & \textbf{97.16\%} & \ul{97.46\%} & \textbf{97.16\%} & \textbf{97.06\%} \\
InceptionResNetV2   & 90.07\% & 87.73\% & 90.07\% & 87.92\% \\
NASNetLarge         & 89.36\% & 90.10\% & 89.36\% & 87.96\% \\
NASNetLarge + GCN & 92.91\% & 93.59\% & 92.91\% & 92.95\% \\
NASNetLarge + GAT & 91.49\% & 91.41\% & 91.49\% & 91.33\% \\
NASNetLarge + GraphSAGE & 91.49\% & 91.57\% & 91.49\% & 91.27\% \\
NASNetMobile        & 91.49\% & 91.40\% & 91.49\% & 91.08\% \\
ResNet101           & 46.10\% & 45.24\% & 46.10\% & 36.43\% \\
ResNet101 + GCN & 62.41\% & 66.52\% & 62.41\% & 56.18\% \\
ResNet101 + GAT & 77.30\% & 74.59\% & 77.30\% & 72.90\% \\
ResNet101 + GraphSAGE & 80.85\% & 79.69\% & 80.85\% & 78.11\% \\
ResNet152           & 57.45\% & 43.31\% & 57.45\% & 48.26\% \\
ResNet152 + GCN & 65.25\% & 57.83\% & 65.25\% & 58.55\% \\
ResNet152 + GAT & 68.79\% & 62.10\% & 68.79\% & 62.66\% \\
ResNet152 + GraphSAGE & 55.32\% & 67.60\% & 55.32\% & 51.35\% \\
ResNet50V2          & 95.04\% & 94.78\% & 95.04\% & 94.30\% \\
ResNet50V2 + GCN & 94.33\% & 95.13\% & 94.33\% & 93.55\% \\
ResNet50V2 + GAT & 95.74\% & 95.96\% & 95.74\% & 95.69\% \\
ResNet50V2 + GraphSAGE & 94.33\% & 94.49\% & 94.33\% & 94.12\% \\
\hline
\end{tabular}
}
\end{table}

\begin{table*}[!ht]
\centering
\caption{Ablation Study of Sequential CNN-GNN Model}
\label{tab:ablation}
\begin{tabular}{lcccc}
    \hline
    \textbf{Model Variant} & \textbf{Accuracy ($\uparrow$)} & \textbf{Precision ($\uparrow$)} & \textbf{Recall ($\uparrow$)} & \textbf{F1 Score ($\uparrow$)} \\
    \hline
    MobileNetV2 w/o GraphSAGE & 93.62\% & 93.75\% & 93.62\% & 92.77\% \\
    GraphSAGE w/o MobileNetV2 & 91.30\% & 90.76\% & 91.30\% & 90.22\% \\
    MobileNetV2 + GraphSAGE (Parallel) & 95.74\% & 96.36\% & 95.74\% & 94.96\% \\
    GNN + MobileNetV2 (Sequential) & 96.45\% & 97.07\% & 96.45\% & 95.67\% \\
    \textbf{Sequential MobileNetV2 + GraphSAGE (Proposed)} & \textbf{97.16\%} & \textbf{97.51\%} & \textbf{97.16\%} & \textbf{96.79\%} \\
    \hline
\end{tabular}
\end{table*}

\begin{table}[ht]
\centering
\caption{CNN and GNN Models and their Parameter Counts}
\label{tab:param}
\begin{tabular}{lc}
\hline
\textbf{Model} & \textbf{Parameter Count ($\downarrow$)} \\
\hline
MobileNetV2          & 2,257,984      \\
EfficientNetB0       & 4,049,571      \\
ResNet50             & 23,587,712     \\
VGG16                & 14,714,688     \\
VGG19                & 20,024,384     \\
Xception             & 20,861,480     \\
DenseNet121          & 7,037,504      \\
DenseNet169          & 12,642,880     \\
DenseNet201          & 18,321,984     \\
InceptionV3          & 21,802,784     \\
NASNetLarge          & 84,916,818     \\
ResNet101            & 42,658,176     \\
ResNet152            & 58,370,944     \\
ResNet50V2           & 23,564,800     \\
GCN                  & 10,666         \\
GAT                  & 25,418         \\
GraphSAGE            & 67,210         \\
\textbf{MobileNetV2+GraphSAGE} & 2,325,194 \\
\hline
\end{tabular}
\end{table}

\subsection{Ablation Studies}\label{ablation}
To better understand the contribution of each component of the proposed model, we conducted an ablation study, the results of which are summarized in Table \ref{tab:ablation}. This study evaluated the performance of different model variants, including MobileNetV2 w/o GraphSAGE, GraphSAGE w/o MobileNetV2, and other combinations.

\begin{enumerate}
\item \textit{MobileNetV2 w/o the GraphSAGE} variant, which uses only MobileNetV2 for feature extraction, achieved an accuracy of 93.62\%.
\item The \textit{GraphSAGE w/o MobileNetV2} variant, which uses only GraphSAGE for feature aggregation, achieved an accuracy of 91.30\%.
\item The \textit{MobileNetV2 + GraphSAGE (parallel)} model, where MobileNetV2 and GraphSAGE are applied in parallel, resulted in an accuracy of 95.74\%.
\item The \textit{GraphSAGE + MobileNetV2 (Sequential)} model, where GraphSAGE is applied after MobileNetV2, achieved an accuracy of 97.16\%.
\end{enumerate}

The proposed sequential MobileNetV2+GraphSAGE model outperformed all other variants, with an accuracy of 97.16\%, demonstrating that the sequential combination of MobileNetV2 and GraphSAGE is the most effective approach for this task. This ablation study highlights the importance of both components working together in a sequential manner rather than independently or in parallel.

\begin{figure}[!ht]
\centering
\includegraphics[width=0.5\textwidth]{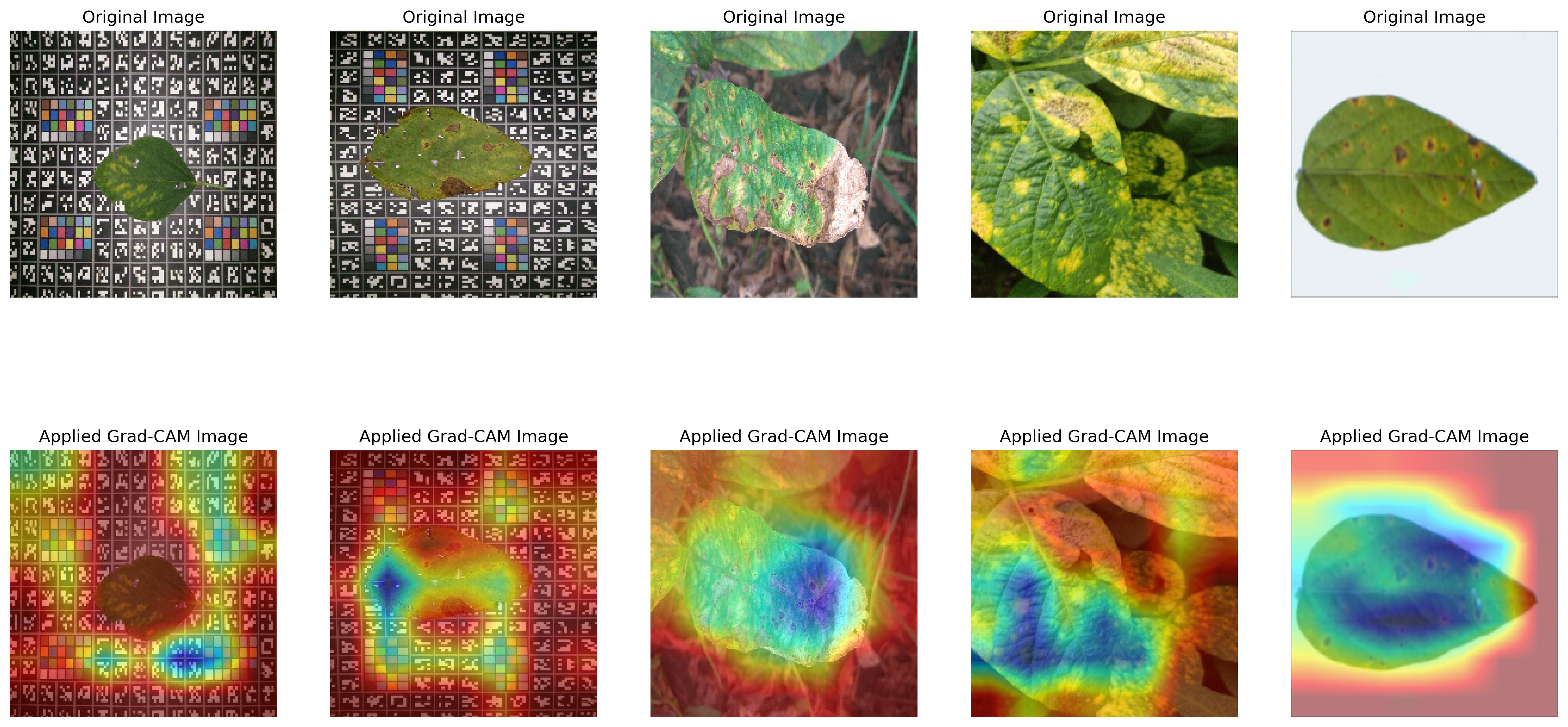}
\label{fig:grad_cam}
\end{figure}

\begin{figure}[!ht]
\centering
\includegraphics[width=0.5\textwidth]{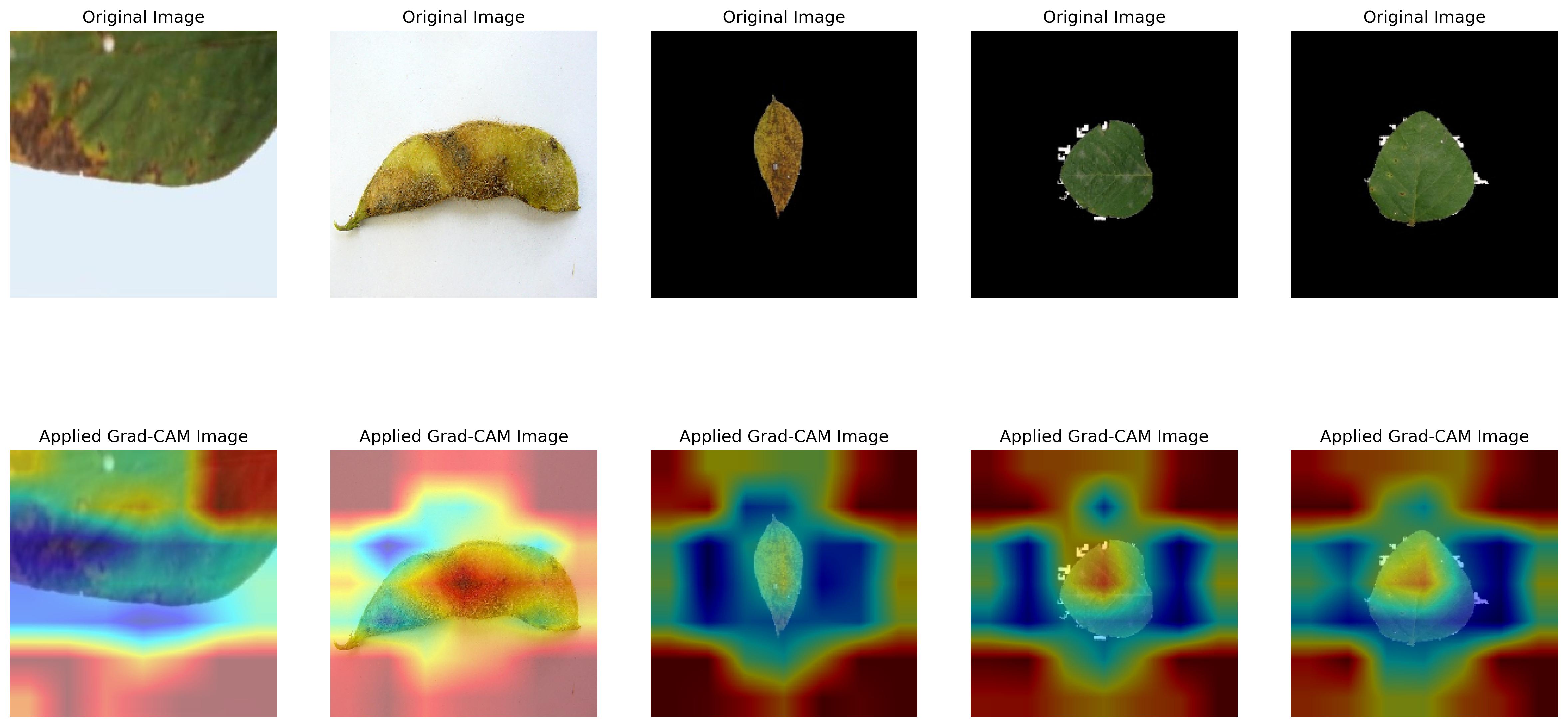}
\caption{Grad-CAM Visualizations. The figure shows the original images (top row) alongside the corresponding Grad-CAM heatmaps (bottom row), highlighting the areas of interest that influence the model's classification decision.}
\label{fig:grad_cam1}
\end{figure}

\begin{figure}[!ht]
\centering
\includegraphics[width=0.5\textwidth]{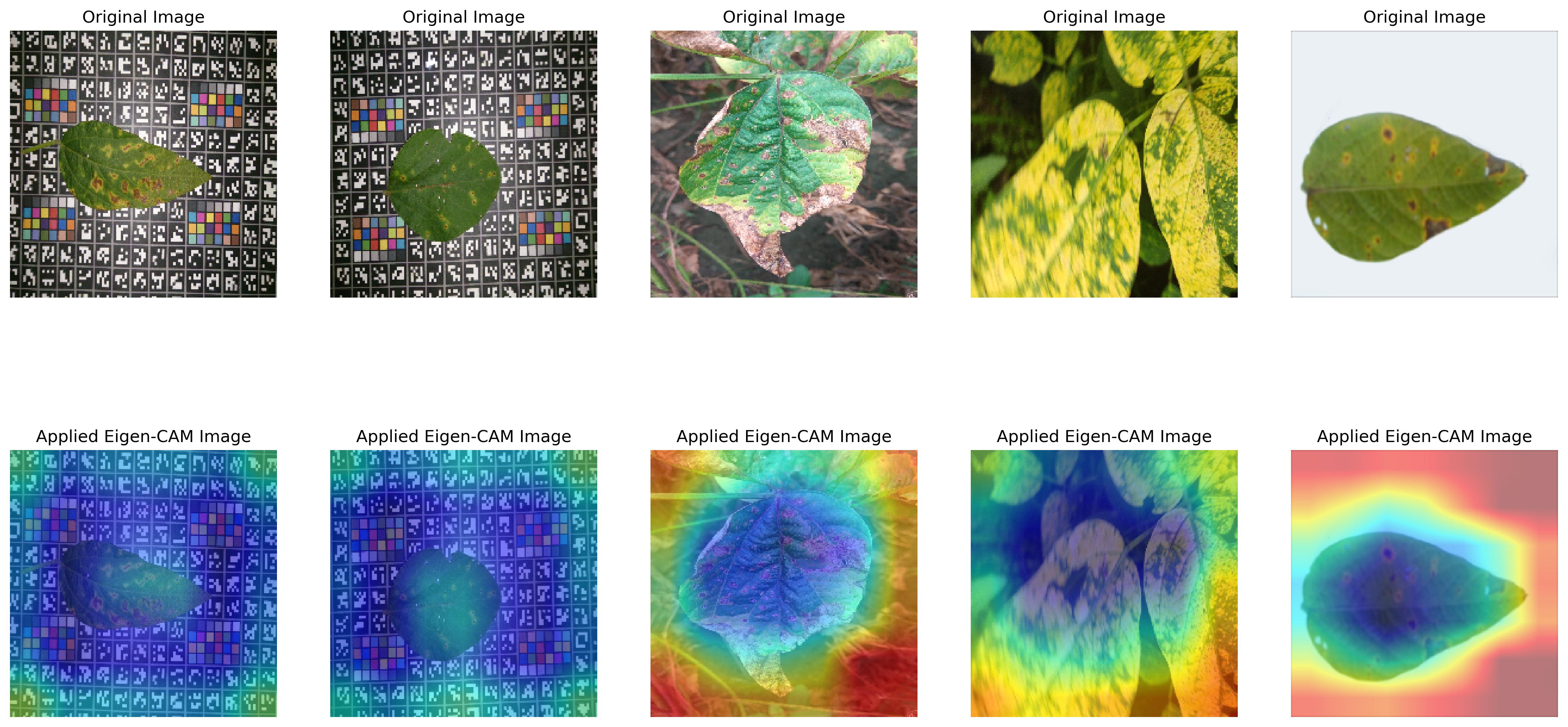}
\label{fig:eigen_cam}
\end{figure}

\begin{figure}[!ht]
\centering
\includegraphics[width=0.5\textwidth]{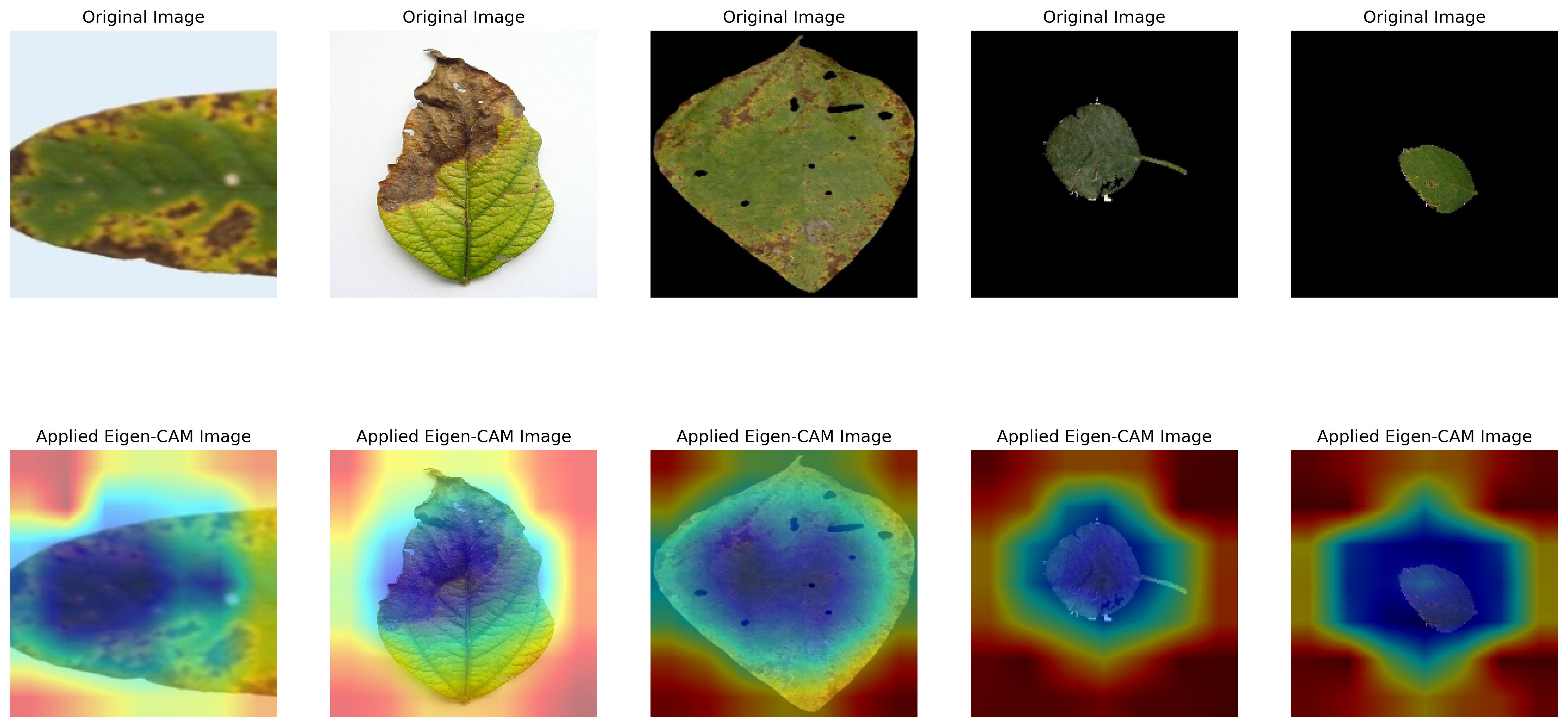}
\caption{Eigen-CAM Visualizations. The figure shows the original images (top row) and the applied Eigen-CAM heatmaps (bottom row), providing a detailed view of how the model interprets different features within the leaf images.}
\label{fig:eigen_cam1}
\end{figure}

\subsection{Interpretability}
The interpretability of the model was evaluated via Grad-CAM and Eigen-CAM to visualize the regions of the images that contribute most to the model's decisions. These techniques allow us to understand which parts of the soybean leaves the model focuses on when classifying diseases.

The Grad-CAM visualizations, shown in Figure \ref{fig:grad_cam1}, indicate that the model focuses primarily on the areas of the leaves that show clear signs of disease, such as lesions, spots, and discoloration. The figure shows the original images (top row) alongside the corresponding Grad-CAM heatmaps (bottom row), highlighting the areas of interest that influence the model's classification decision. The Eigen-CAM visualizations, shown in Figure \ref{fig:eigen_cam1}, further help identify the key features associated with each disease type, confirming that the model learns discriminative features essential for accurate classification. The Eigen-CAM heatmaps provide a detailed view of how the model interprets different features within the leaf images. These visualizations provide confidence in the model's decision-making process and demonstrate its ability to focus on relevant image patterns, making it more interpretable and trustworthy.

\section{Conclusion and Future Work}\label{sec6}
This work presented a hybrid soybean leaf disease detection framework that combines MobileNetV2 for efficient feature extraction and GraphSAGE for capturing symptom relationships. The resulting MobileNetV2 + GraphSAGE model achieved 97.16\% accuracy, 97.51\% precision, and a 96.79\% F1 score, outperforming alternative architectures while maintaining low computational cost and fast inference, which are the key factors for deployment in mobile and edge environments. Grad-CAM and Eigen-CAM provided further insights into the visual and relational features driving predictions, enhancing the model’s interpretability and trustworthiness for real-world use. This approach underscores the potential of integrating GNNs with lightweight CNNs for efficient, scalable plant disease detection.

Future work will focus on improving generalization through expanded datasets incorporating environmental variables (e.g., soil and climate), optimizing the model by pruning and quantization, and investigating advanced architectures such as attention-based and multi-modal models. Using TensorFlow Lite and ONNX deployment strategies ensures real-time performance on resource-constrained devices, enabling practical smart farming solutions.

\bibliographystyle{unsrt}
\bibliography{main}

\begin{IEEEbiography}[{\includegraphics[width=1in,height=1.25in,clip,keepaspectratio]{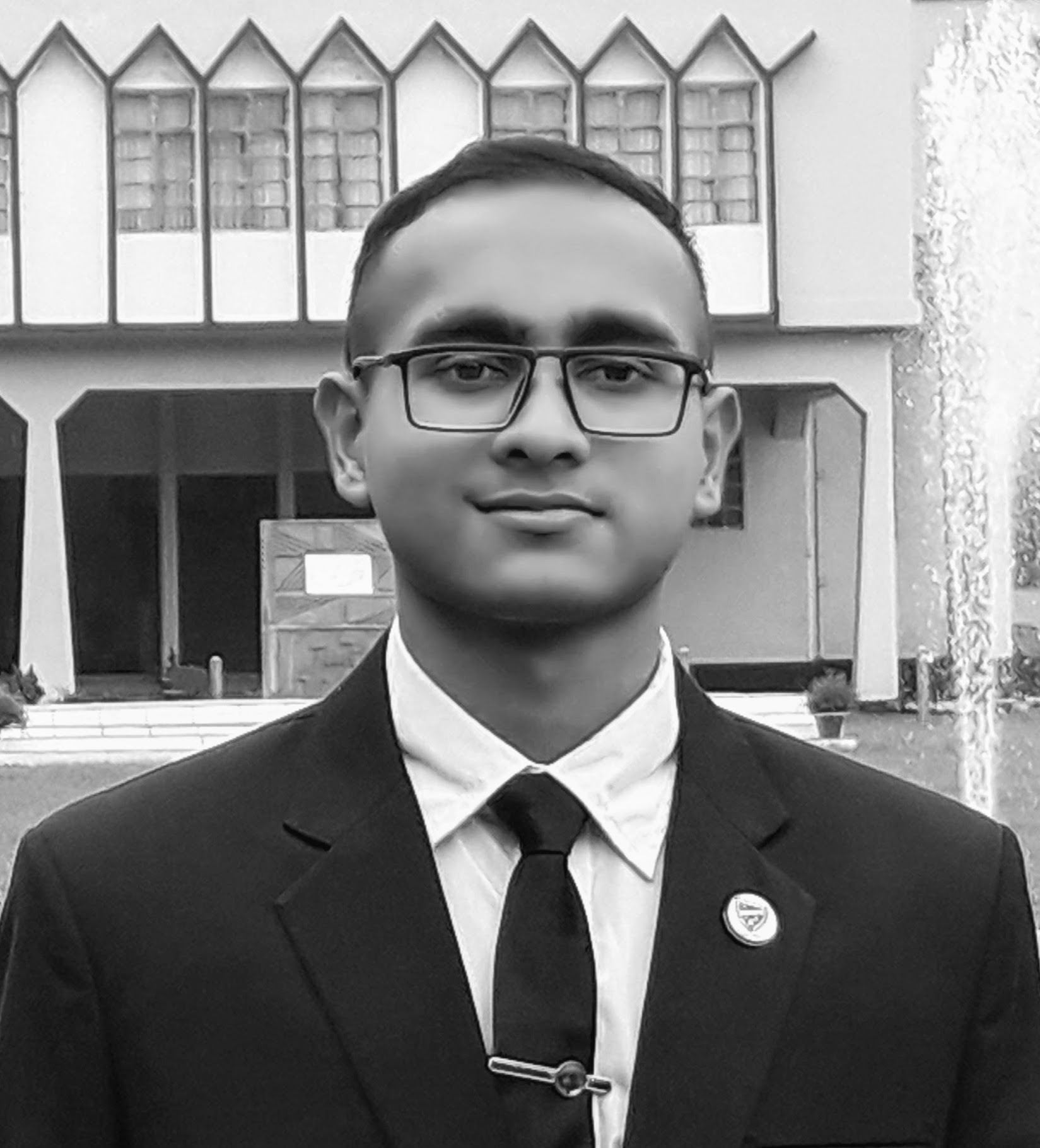}}]{Md Abrar Jahin} is a Ph.D. student and Graduate Fellow with the Thomas Lord Department of Computer Science, Viterbi School of Engineering, University of Southern California (USC), Los Angeles, CA, USA. He is also affiliated as an AI Researcher and Multidisciplinary Scientist with the Center on Knowledge Graphs, Information Sciences Institute (ISI), located in Silicon Beach, CA, USA. He received the B.Sc. degree in Industrial \& Production Engineering from the Department of Industrial Engineering and Management, Khulna University of Engineering and Technology (KUET), Khulna, Bangladesh, in March 2024. From March 2024 to March 2025, he served as a Visiting Researcher at the Okinawa Institute of Science and Technology Graduate University (OIST), Japan, and as a Lead Researcher at the Advanced Machine Intelligence Research (AMIR) Lab, Bangladesh.

His current research interests include efficient deep learning, quantum machine learning, geometric deep learning, and trustworthy artificial intelligence, with applications in high-energy physics, healthcare, and supply chain optimization. His previous research contributions span reinforcement learning, sentiment analysis, operations research, and comparative genomics. Mr. Jahin was the inaugural recipient of the Student Researcher of the Year Award 2024 from the KUET Research Society for publishing the highest number of high-impact research articles and demonstrating exceptional leadership between October 2023 and November 2024. More information is available at: \href{https://abrar2652.github.io/}{https://abrar2652.github.io/}.
\end{IEEEbiography}

\begin{IEEEbiography}[{\includegraphics[width=1in,height=1.25in,clip,keepaspectratio]{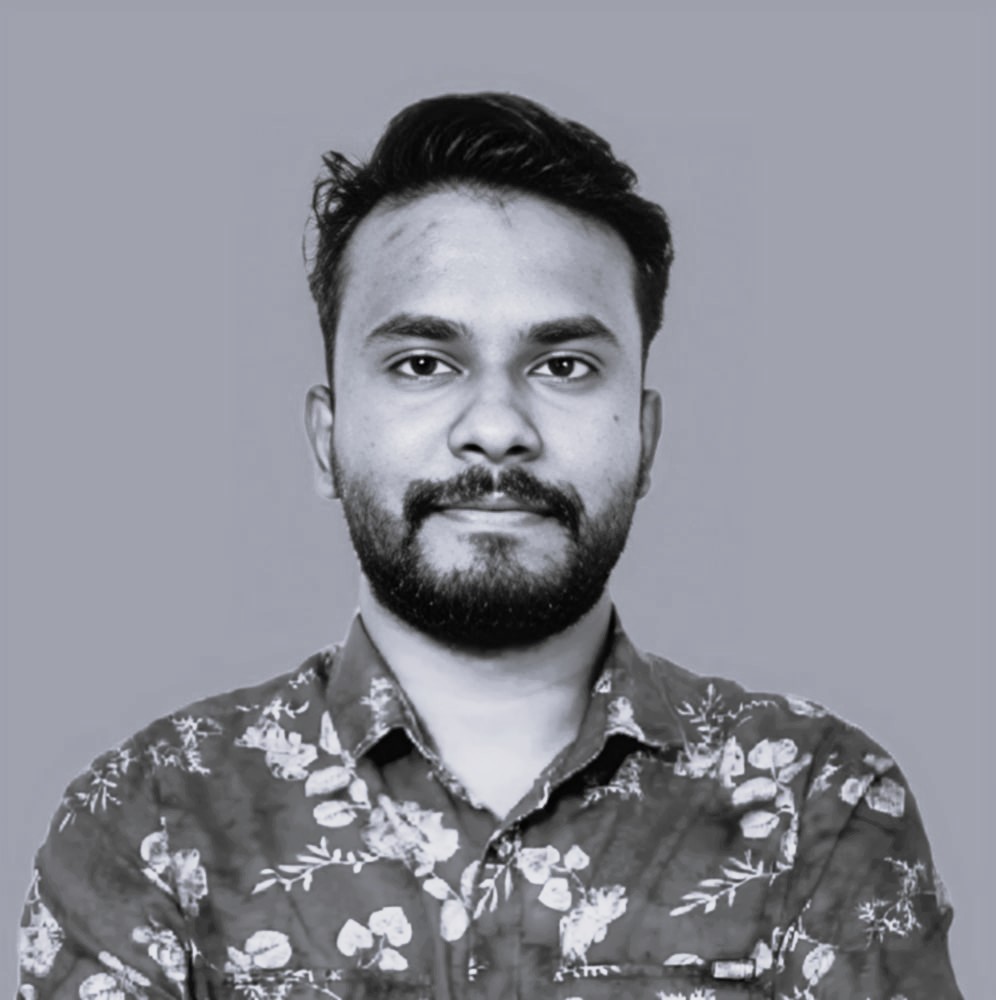}}]{Shahriar Soudeep} received his B.Sc. in Computer Science \& Engineering (2024) from the American International University-Bangladesh (AIUB), Dhaka, Bangladesh. He is currently a Research Assistant at the Advanced Machine Intelligence Research Lab (AMIRL), Bangladesh. His research interests include: machine learning, deep learning, graph neural networks, and computer vision. His work on transformer models has been published in \textit{Sustainable Cities and Society}, Elsevier.
\end{IEEEbiography}

\begin{IEEEbiography}[{\includegraphics[width=1in,height=1.25in,clip,keepaspectratio]{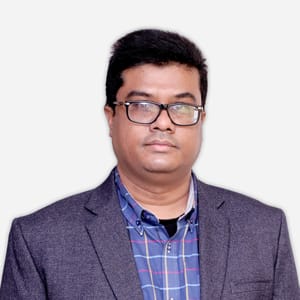}}]{M. F. Mridha (Senior Member IEEE, Professional Member ACM)} is currently working as a Professor in the Department of Computer Science, American International University-Bangladesh (AIUB). He also worked as Associate Professor and Chairman in the Department of Computer Science and Engineering, Bangladesh University of Business and Technology (BUBT) from 2019 to 2022 as a CSE department faculty member at the University of Asia Pacific, and as a graduate head from 2012 to 2019. He is the founder and director of the Advanced Machine Intelligence Research Lab (AMIR Lab). He received his Ph.D. in the domain of AI from Jahangirnagar University in 2017. 

For more than 20 (Twenty) years, he has been with the master’s and undergraduate students as a supervisor of their thesis work. He has authored/edited several books with Springer and published more than 250 Journal and Conference papers. His research interests include Artificial Intelligence (AI), Computer Vision (CV), Machine Learning(ML), Deep Learning(DL), and Natural Language Processing (NLP), etc. He has served as a program committee member in several international conferences/workshops. He served as an Editorial Board Member of several journals, including the PLOS ONE Journal. He was among the top 2\% of scientists worldwide in the 2024 edition of the Stanford University/Elsevier. He achieved the top 5 most productive researchers in Bangladesh by (Scopus/ Elsevier) and secured the Top researcher in the field of Computer Science and Engineering in 2024.
\end{IEEEbiography}

\begin{IEEEbiography}[{\includegraphics[width=1in,height=1.25in,clip,keepaspectratio]{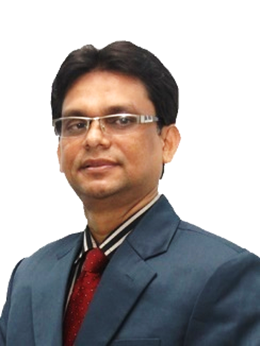}}]{Ts. Dr. Md. Jakir Hossen (Senior Member, IEEE)} is currently working as an Associate Professor in the Department of Robotics and Automation, Faculty of Engineering and Technology, Multimedia University, Melaka, Malaysia. He received a master’s degree in communication and network engineering from Universiti Putra Malaysia, Malaysia, in 2003. He received the PhD degree in smart technology and robotic engineering from Universiti Putra Malaysia, Malaysia, in 2012. His research interests are the applications of artificial intelligence techniques in data analytics, robotics control, data classifications, and predictions. He can be contacted at email: jakir.hossen@mmu.edu.my. 
\end{IEEEbiography}

\begin{IEEEbiography}[{\includegraphics[width=1in,height=1.25in,clip,keepaspectratio]{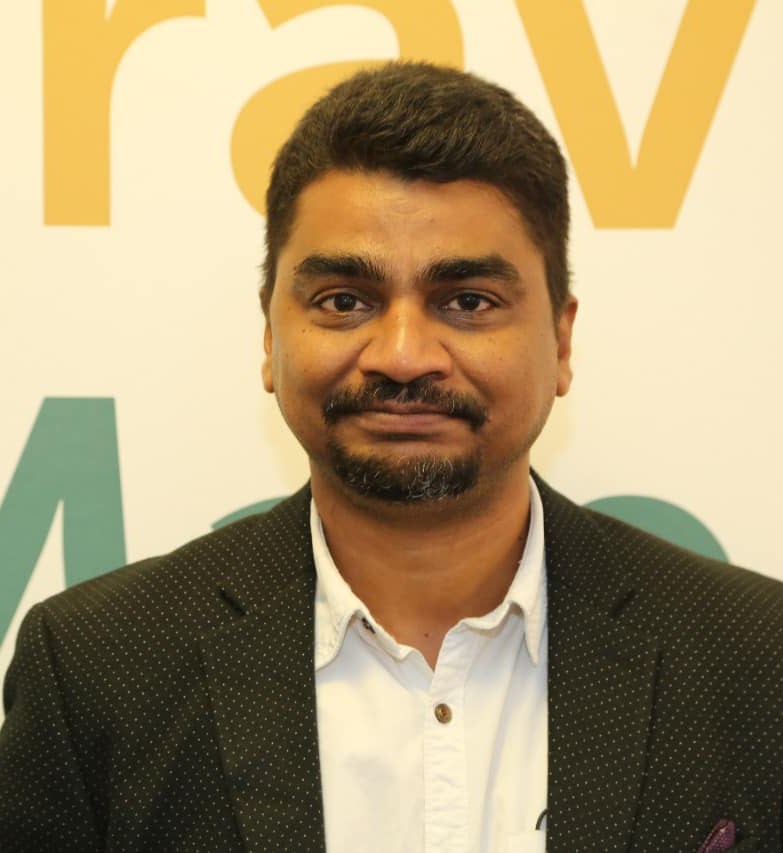}}]{Nilanjan Dey (Senior Member, IEEE)} received the B.Tech., M.Tech. in information technology from West Bengal Board of Technical University and Ph.D. degrees in electronics and telecommunication engineering from Jadavpur University, Kolkata, India, in 2005, 2011, and 2015, respectively. Currently, he is an Associate Professor with the Techno International New Town, Kolkata, and a visiting fellow of the University of Reading, UK. He has authored over 300 research articles in peer-reviewed journals and international conferences, and 40 authored books. His research interests include medical imaging and machine learning. Moreover, he actively participates in program and organizing committees for prestigious international conferences, including World Conference on Smart Trends in Systems Security and Sustainability (WorldS4), International Congress on Information and Communication Technology (ICICT), International Conference on Information and Communications Technology for Sustainable Development (ICT4SD), etc.

He is also the Editor-in-Chief of the International Journal of Ambient Computing and Intelligence, Associate Editor of IEEE Transactions on Technology and Society, and series Co-Editor of Springer Tracts in Nature-Inspired Computing and Data-Intensive Research from Springer Nature and Advances in Ubiquitous Sensing Applications for Healthcare from Elsevier, etc. Furthermore,  he was an Editorial Board Member of Complex \& Intelligence Systems, Springer, Applied Soft Computing, Elsevier and he is an Editorial Board Member of International Journal of Information Technology, Springer, International Journal of Information and Decision Sciences, etc. He is a Fellow of IETE and a member of IE, ISOC, etc.
\end{IEEEbiography}

\EOD

\end{document}